\documentclass[10pt,twocolumn,letterpaper]{article}

\usepackage{cvpr}      

\usepackage{graphicx}
\usepackage{amsmath}
\usepackage{amssymb}
\usepackage{booktabs}
\usepackage{epsfig}
\usepackage{indentfirst}
\usepackage{multirow}
\usepackage[labelfont=bf]{caption}

\usepackage[pagebackref,breaklinks,colorlinks]{hyperref}

\usepackage[capitalize]{cleveref}
\crefname{section}{Sec.}{Secs.}
\Crefname{section}{Section}{Sections}
\Crefname{table}{Table}{Tables}
\crefname{table}{Tab.}{Tabs.}

\def\cvprPaperID{2035} 
\def\confName{CVPR}
\def\confYear{2022}

\begin{document}

\title{Generating Unrestricted 3D Adversarial Point Clouds}

\author{Xuelong Dai$^{1}$\thanks{$\dagger$ Corresponding author}, Yanjie Li$^{1}$,  Hua Dai$^{2}$,  Bin Xiao$^{1}$\\
$^{1}$The Hong Kong Polytechnic University, \quad $^{2}$Nanjing University of Posts and Telecommunications \\
{\tt\small\{xuelong.dai,yanjie.li\}@connect.polyu.hk};
{\tt\small daihua@njupt.edu.cn};\\
{\tt\small csbxiao@comp.polyu.edu.hk}
}

\maketitle

\begin{abstract}
   Utilizing 3D point cloud data has become an urgent need for the deployment of artificial intelligence in many areas like facial recognition
   and self-driving. However, deep learning for 3D point clouds is still vulnerable to adversarial attacks, e.g., iterative attacks, point transformation attacks, and generative attacks. These attacks need to restrict perturbations of adversarial examples within a strict bound, leading to the unrealistic adversarial 3D point clouds.
   In this paper, we propose an \underline{Adv}ersarial \underline{G}raph-\underline{C}onvolutional \underline{G}enerative \underline{A}dversarial \underline{N}etwork (AdvGCGAN) to generate visually realistic adversarial 3D point clouds from scratch. 
   Specifically, we use a graph convolutional generator and a discriminator with an auxiliary classifier to generate realistic point clouds, which learn the latent distribution from the real 3D data. The unrestricted adversarial attack loss is incorporated in the special adversarial training of GAN, which enables the generator to generate the adversarial examples to spoof the target network. Compared with the existing state-of-art attack methods, the experiment results demonstrate the effectiveness of our unrestricted adversarial attack methods with a higher attack success rate and visual quality. Additionally, the proposed AdvGCGAN can achieve better performance against defense models and better transferability than existing attack methods with strong camouflage.

\end{abstract}

\section{Introduction}

Deep learning has been widely applied in many scenarios like smart city \cite{chen2019survey,kok2017deep}, healthcare \cite{esteva2017dermatologist}, automotive \cite{kendall2019learning, Luo_2021_CVPR}, etc.
Its ability is proved to be qualified for the tasks that even humans are not competent, especially in computer vision tasks \cite{krizhevsky2012imagenet, he2016deep, redmon2016you}.
In recent years, 3D deep learning has made a huge improvement in various 3D tasks, for example, 3D classification \cite{qi2017pointnet, li2018so} and 3D segmentation \cite{engelmann2017exploring, landrieu2019point, landrieu2018large}, which make real-world applications like facial recognition \cite{gilani2018learning} and self-driving \cite{li2019gs3d} achieve better performance than only using 2D models. 

While rapid improvement has been achieved in 3D deep learning models, 3D deep learning is also vulnerable to adversarial attacks just like 2D models \cite{xiang2019generating, zeng2019adversarial}. Adversarial robustness is rather important because it can cause many security problems in real-world applications. Many recent works \cite{dong2020self, liu2019extending, zhao2020isometry, wicker2019robustness, zhou2020lg} show that even with few points modification, the state-of-art 3D deep learning models will misclassify.
 
\begin{figure}[t] 
\setlength{\abovecaptionskip}{-0.2cm}   

\setlength{\belowcaptionskip}{-0.5cm} 
   \begin{center}
     \includegraphics[width=0.8\linewidth]{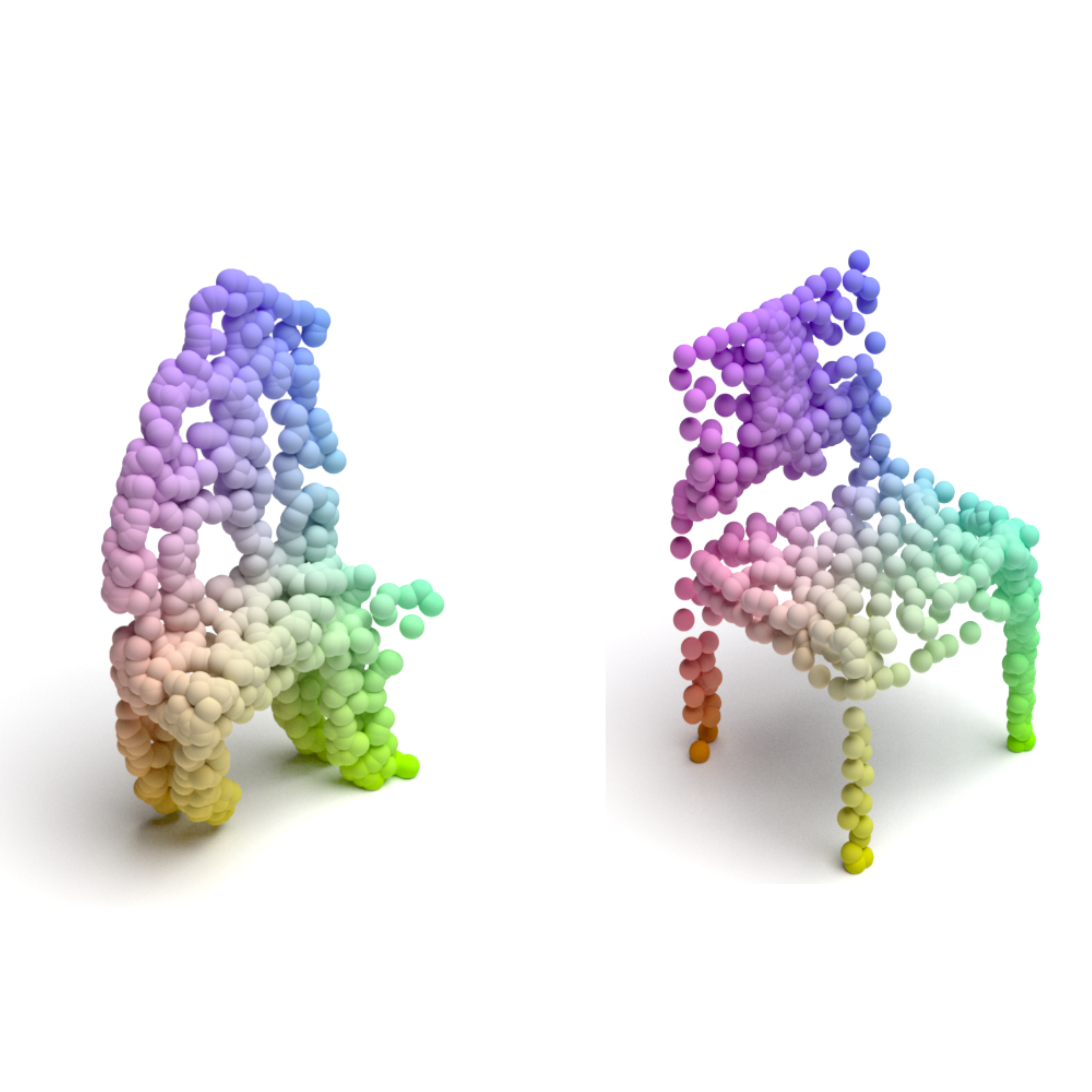}
   \end{center}
      \caption{\textbf{An example of the adversarial examples generated by the state of art perturbation-based adversarial attack method (left)}. The deformation problem occurs in the perturbation-based adversarial examples, while the unrestricted adversarial example generated by our AdvGCGAN is visually more realistic (right).}
   \label{fig:example}
   \end{figure}

Recent works in 3D adversarial attack methods can be categorized into 2D-attack-based methods like FGSM \cite{liu2019extending} and C\&W \cite{xiang2019generating} in the 3D scenario, point-based methods like point detaching \cite{wicker2019robustness} and isometry transformation \cite{zhao2020isometry}, generative-based methods like LG-GAN \cite{zhou2020lg}. All of these methods have weaknesses in some aspects. For example, directly using FGSM and C\&W in the 3D scenario will generate additional outlier points and are very time-consuming. The generative-based methods generate very unnatural point clouds, which results in bad performance against defense methods. Moreover, nearly all of the previous works do not adopt the local features from graph-convolutional operations, which further limits the visual quality of the adversarial examples.

The unrealistic adversarial example problem is one of the most crucial problems that remain to be solved. They should be \textit{indistinguishable to human eyes} according to the original definition of adversarial examples \cite{goodfellow2014explaining}. In practice, only the adversarial examples that deceive both humans and deep learning models can lead to severe security problems. These unrealistic adversarial examples are also hard to deploy in the physical world, e.g., the outlier points do not physically exist in a real-world object. In addition, most of the unrealistic adversarial examples can be defended by different defense mechanisms.

To solve the mentioned problems, we propose the unrestricted adversarial examples generation algorithm for 3D point clouds. Our method aims to generate natural adversarial point clouds and at the same time guarantee a high attack success rate. As shown in Figure 1, our unrestricted adversarial example (right) is visually more realistic than the point cloud generated by the previous study. We verified the visual quality by adopting the metrics from the traditional 3D GAN task \cite{achlioptas2018learning}. The norm or distance restrictions adopted in the previous study can not fully represent the differences between two point clouds. Shape and local features are more important than the distance between points. Therefore, we generate adversarial examples from scratch with a GAN instead of adding perturbations over the existing data.

More specifically, in this paper, we propose a novel adversarial graph-convolutional generative adversarial network (AdvGCGAN). We adopt the GAN architecture for the realistic point cloud generation. We first train the AdvGCGAN with GAN traning for standard GAN task. Then, to achieve adversarial attack, we propose unrestricted adversarial attack loss and adversarial training to enable the AdvGCGAN to conduct attacks aginst the target network.
When generating the adversarial examples, the network will search the latent space between different classes to find the examples that can lead to misclassification of the target network. To support multi-class generation, we add the label information into the training of the generator and an auxiliary classifier into the discriminator. These modifications aid the search process and ensures that the network generates the meaningful point clouds. 
The adversarial examples generated with the above procedures are called unrestricted adversarial examples. The generation process is not bounded by a specific norm, which leads to natural and realistic adversarial examples. The detailed network structure will be given in the later section.

We experiment on different state-of-art 3D point cloud classification networks (PointNet, PointNet++, and DGCNN) to demonstrate the effectiveness of our method. The experiment results show that our method can generate adversarial examples with a high attack success rate and still can deceive human eyes.

The contributions of our work are summarized as follows:

\begin{itemize}
\setlength{\itemsep}{0pt}

\setlength{\parsep}{0pt}

\setlength{\parskip}{0pt}
   \item We are the first to propose a general 3D unrestricted adversarial examples generation method against 3D point cloud deep learning networks, which solves the unrealistic adversarial examples generation problem caused by the current perturbation
   -based attack methods.
   \item We present a graph convolutional GAN-based network design for 3D adversarial examples generation. The label information is also considered in the proposed AdvGCGAN. a special two-stages training method and the unrestricted adversarial attack loss are proposed for effective and stable training.
   \item Experiments on the different 3D deep learning classification networks demonstrate the effectiveness of our proposed method. Comparison with the state-of-art existing attack methods proved that our method can achieve a higher success rate against different defense methods, and the adversarial examples still achieve high performance on 3D GAN evaluation metrics. 
\end{itemize}

\section{Related Work}

\subsection{3D deep learning}

Before the appearance of deep learning, the feature extraction in 3D datasets was a difficult task. 3D data has richer information than 2D images. There are also different types of 3D data, like voxel, mesh, point cloud, etc. 
All of these problems limited the development of 3D feature learning. As the emergence of deep learning, PointNet \cite{qi2017pointnet} was the first deep learning network that remarkably improves the performance of machine learning for the 3D point cloud dataset.
PointNet achieved significant accuracy on both 3D object classification and 3D segmentation tasks. It adopted a symmetric function to extract the feature with unordered input. Since the success of PointNet, there is a surge of researches in 3D deep learning.
PointNet++ \cite{qi2017pointnet++} further improved the performance of PoinNet by learning more localized structure information. DGCNN \cite{wang2019dynamic} extracted the feature of 3D point clouds by considering a local neighborhood graph. LDGCNN \cite{zhang2019linked} linked the hierarchical features from dynamic graphs
and increased the performance of LDGCNN. To evaluate the performance of our proposed attack method, we will attack the state-of-art methods to demonstrate our method's effectiveness.

\subsection{3D adversarial attacks}

While 3D deep learning has made great success, the robustness of these methods has also drawn researchers' attention. 
As we mentioned before, we classify the existing adversarial attack algorithm into three categories:

\textbf{2D-attack-based methods}: The 2D-attack-based methods can further be categorized into the gradient-based attack and optimization-based attack.
    The FGSM and IFGM proposed in \cite{liu2019extending} adopted the gradient of the 3D deep learning network to find the perturbation that will fool the target network.
    Zheng et al. \cite{zheng2019pointcloud} proposed a point dropping attack that drops the point that has the lowest salience scores, which is calculated from a gradient-based saliency map.
    Xiang et al. \cite{xiang2019generating} transferred the famous C\&W from the 2D scenario into the 3D point cloud and proposed an optimization-based attack algorithm to find the minimal perturbation
    that can cause the misclassification of the target model. However, these methods are also easy to defense by adversarial training (and other 2D defense methods). 
    Also, their methods generate many outlier points outside the adversarial point clouds, which can be visually identified by human eyes.

\textbf{Point-based methods}: The point-based attack methods consider the characteristic of the 3D dataset. 
The data is the coordinate of points rather than pixels. Thus, these methods focus on performing attacks by point occlusion, 
point attach/detach, and rotations, which is very effective in the 3D dataset. Wicker et al. \cite{wicker2019robustness} proposed a point detach algorithm that iteratively removes the 
point from PointNet's critical point set, and successfully deceives the PointNet. Zhang et al. \cite{zhang2019adversarial} proposed a point attach algorithm to attach an adversarial cluster to fool the target network.
These methods also mostly aim to attack the PointNet network, which is very vulnerable to point removal. The transferability of these attacks is limited.
With a better network like DGCNN, their performance will drop severely. Zhao et al. \cite{zhao2020isometry} proposed an isometry transformation attack that can easily fool the 3D deep learning network with simple rotations. However, their attack can be easily defended by 
data augmentation. 

\textbf{Generative-based methods}: The generative-based attack methods use VAE or GAN network to generate adversarial examples. 
Hamdi et al. proposed an Auto-Encoder based attack model called AdvPC \cite{hamdi2020advpc}. Their model aimed to propose more transferable attack perturbations while can break through the 3D deep learning defense methods. 
 Zhou et al. \cite{zhou2020lg} proposed a label-guided GAN-based attack model for the 3D deep learning network. Their work takes the clean point clouds as input and outputs the adversarial examples after an Encoder-Decoder generator. 
They also add label information for the targeted attack. Zhang et al. \cite{zhang2021td} performed an adversarial attack using an Encoder-Decoder network. Their adversarial examples are generated by limiting the $\mathit{Chamfer}$ distance measurement.
These generative-based methods take the real 3D point cloud data as input and output the adversarial examples to fool the given 3D network.
 However, these methods also have meaningless points generated like outlier points, which are easy to identify by human eyes. In this paper, we propose our unrestricted adversarial example generation method, which directly generates adversarial examples from scratch without adding restrictions over distance. 

\section{Unrestricted 3D Adversarial Attacks}

In this section, we will give the detailed network architecture of AdvGCGAN. We first illustrate the definition of 
unrestricted 3D adversarial examples. Then, the generation of these examples is implemented by our
AdvGCGAN. The structures of the graph-convolutional generator network and discriminator network 
are explained in the following.
\subsection{Unrestricted 3D Adversarial Examples}

Unlike the perturbation-based methods, the unrestricted adversarial examples are generated without any
norm or distance restrictions. Motivated by the definition of Song's work \cite{song2018constructing} in the 2D scenario, we give the definition of 3D unrestricted adversarial examples.
Suggest o : $\mathit{O} $ takes 3D data in its domain $O$ and outputs one of $M$ labels $\{1, 2, ..., M\}$. Consider any 3D classifier $f$ also takes 3D data as input, and it outputs its prediction $f(x) \rightarrow \{1, 2, ..., M\}$ by taking the 3D data input $x$. 

\textbf{Definition 1 Restricted 3D Adversarial Examples}. A restricted 3D adversarial example $A_r$ is any 3D data  (3D point cloud in this paper) that fools the classifier $f$ by adding perturbation over a given set of data $D$, i.e.,
\begin{multline}
       A_r \triangleq \{x \in O| \exists  x' \subset D, x=x'+\delta \\ \wedge o(x)\ne f(x)  \wedge f(x)=f(x')=o(x')\}
\end{multline}

\textbf{Definition 2 Unrestricted 3D Adversarial Examples}. An unrestricted 3D adversarial example $A_u$ is any 3D point cloud that makes the classifier $f$ give the wrong prediction, i.e.,
\begin{equation}
   A_u \triangleq \{x \in O|o(x) \neq f(x)\}
\end{equation}
where $\delta$ is the norm or distance restriction to generate adversarial examples from $D$.

Due to the drawback of the unrealistic 3D adversarial point clouds generated by the previous work, we propose the unrestricted adversarial example generation method AdvGCGAN that aims for a more natural point cloud generation.
Limited by the norm or distance restrictions like $l_2$ norm and $\mathit{Chamfer}$ distance, the previous works' generation ability is constrained. Also, these works generate the example based on the clean data, which 
further limits the search space of the generator. Thus, our method removes any restrictions on norm or distance measurements. 

\begin{figure*}
\setlength{\abovecaptionskip}{-0.2cm}   

\setlength{\belowcaptionskip}{-0.5cm} 
   \begin{center} 
   \includegraphics[width=6.8in]{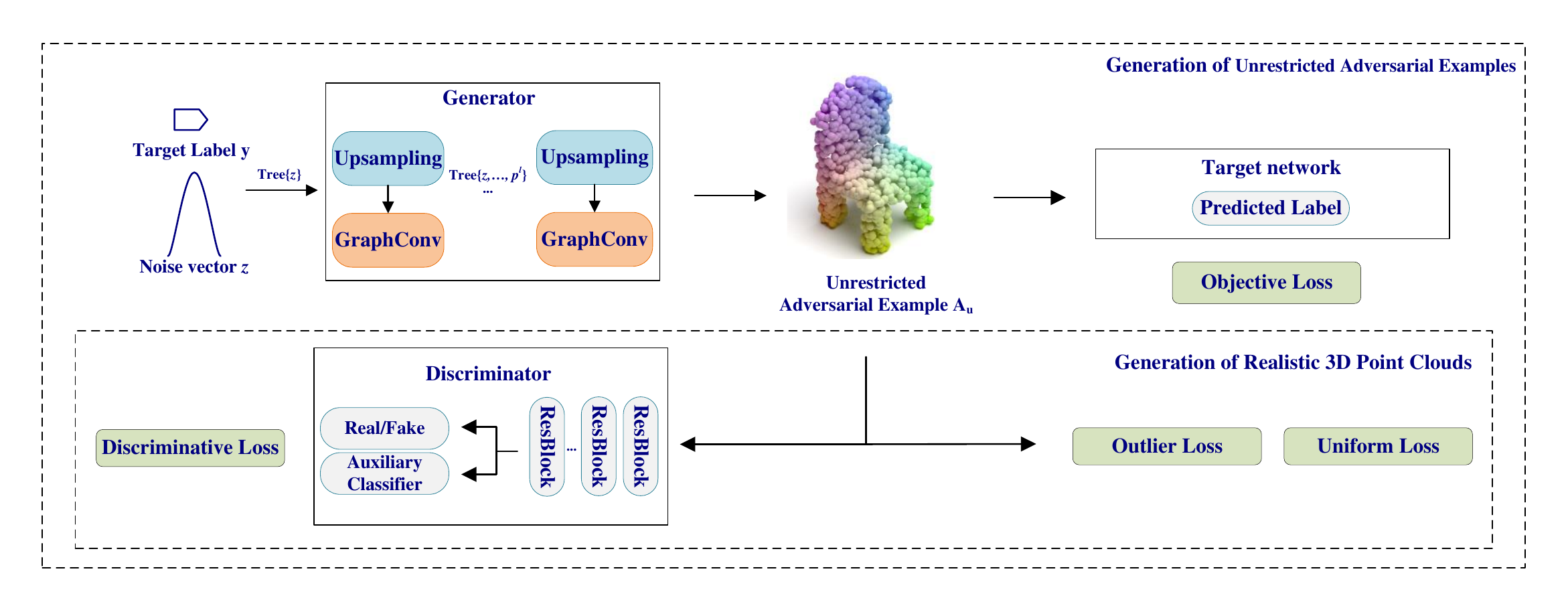}
   \end{center}
      \caption{\textbf{The framework of the proposed AdvGCGAN.} Take the noise vector $z$ and the target label $y$ as the input, the AdvGCGAN generates the corresponding point cloud by its tree structure generator. At each layer of the generator, the Upsampling and GraphConv functions are performed to generate the next set of points. The Upsampling function generates new points by the former generated points stored in the tree, and the GraphConv function aggregates features from the point set. After the final layer of the generator, the 
      unrestricted adversarial example $A_u$ is outputted. To achieve the generation of realistic 3D point clouds, the discriminator differentiates between real and generated point clouds and classifies the point clouds with the auxiliary classifier. Additionally, the outlier loss and the uniform loss forces the generator to preserve the shape and local features of the generated point cloud. The objective loss is utilized for the generation of unrestricted adversarial examples. According to the prediction of $A_u$ given by the targeted network, the objective loss guides the generator to search the latent space to generate the unrestricted adversarial examples.}
   \label{fig:framework}
   \end{figure*}
   
To achieve unrestricted 3D adversarial examples generation, we first aim to train a generative model $g_{G} (z,y)$ that model a set of 3D point cloud data $\mathit{O}$ with a discriminator. The model generate 3D point clouds by taking the random noise $z \in \mathcal{R}^{n}$ and the label $y \in \{1, 2, ..., M\}$ as input. Ideally, we will get $\mathit{O} \equiv \{g_{G} (z,y)| z \in \mathcal{R}^{n}, y \in \{1, 2, ..., M\}\}$. Additionally, the auxiliary classifier $f_{ac}$ in the discriminator should satisfy $y=o(g_{G} (z,y))=f_{ac}(g_{G} (z,y))$.
Then, we adopt adversarial training to $g_{G} (z,y)$ for conducting adversarial attack against the target classifier $f$, i.e., $y=o(g_{G} (z,y))\neq f(g_{G} (z,y)) \wedge o(g_{G} (z,y))=f_{ac}(g_{G} (z,y))$, which satisfies the definition of unrestricted 3D adversarial examples.

In practice, we select AC-GAN \cite{odena2017conditional} structure for the generative model and the discriminator with an auxiliary classifier. To achieve the generation of realistic unrestricted adversarial examples, we will illustrate the details of the proposed adversarial graph-convolutional generative adversarial network (AdvGCGAN) in the following part.

\subsection{Graph-Convolutional GAN Network}

The framework of the proposed AdvGCGAN is given in Figure 2. The AdvGCGAN consists of two networks: a generative network for generating realistic unrestricted adversarial point clouds and a discriminator network with an auxiliary classifier for the better generative ability of the generative network.
\vspace{-1em}
\subsubsection{Network Architecture}
The network architecture of AdvGCGAN is a standard GAN network with a generator and a discriminator. We choose the graph-convolutional generator of TreeGCN \cite{shu20193d} as our unrestricted 3D adversarial generator for its good capability of generating high-quality 3D point clouds. Each layer of the generator contains two operations, GraphConv and Upsampling. The generator will output the generated point cloud after the last operation. To achieve adversarial examples generation, the generator should learn the label information from the input. Therefore, we add label information to the input of the generator, which supports multi-class point cloud generation.

For discriminator, we propose a 3D discriminator with an auxiliary classifier for generating a natural multi-class adversarial point cloud. The discriminator should distinguish clean point clouds from generated adversarial examples, and it should also classify the input point clouds into the right category. The auxiliary classifier is crucial for the unrestricted adversarial example generation because it guides the generator to learn the label feature from the point clouds. To improve the training stability, we also adopt residual blocks and spectral normalization to the discriminator architecture. The detailed design of the discriminator is given in the supplementary materials.

%
\subsubsection{Unrestricted Adversarial Attack Loss}
To achieve realistic unrestricted adversarial examples generation, the generator should first fool the targeted network. Secondly, the generator should also generate a realistic point cloud that still can be correctly classified by the auxiliary classifier. With a random noise $z$, the generator is trained by minimizing the unrestricted adversarial attack loss functions given as follows:
\begin{equation}
\setlength{\abovedisplayskip}{3pt}
\setlength{\belowdisplayskip}{3pt}
   \mathcal{L} = \mathcal{L}_{dis} + \lambda_1\mathcal{L}_{obj}+\lambda_2\mathcal{L}_{out}+\lambda_3\mathcal{L}_{UL} 
\end{equation}
where $\lambda_1$, $\lambda_2$ and $\lambda_3$ are the weight factors. The loss function of the generator contains four parts: the discriminative loss, the objective loss, the outlier loss, and the uniform loss.

For the targeted attack,
\begin{multline}
   \mathcal{L}_{obj}= - Softmaxloss \\ f(y_{target} | g_{G} (z,y_{t}) :y_f = y_{t})
\end{multline}
where $g_{G}$ is the generator, $f$ is the target network, $y_{target}$ is the target label for the targeted attack, $y_f$ is the label predicted by the target network, $y_{t}$ is the true label for generating the unrestricted adversarial examples, and $Softmaxloss$ is the combination of a cross-entropy loss and a softmax function \cite{krizhevsky2012imagenet}. 

The objective loss sums the loss from the correctly classified generated point clouds, the objective loss guides the generator to generate point clouds that make $f$ predict $y_{target}$.

For the untargeted attack,
\begin{multline}
   \mathcal{L}_{obj}=- Softmaxloss \\ \max _{y\neq y_{t}}  f(y | g_{G} (z,y_{t}) :y_f = y_{t} )
\end{multline}

The objective loss guides the generator to generate point clouds that make $f$ predict the second-largest classification label.

For the discriminative loss,
\begin{multline} 
   \mathcal{L}_{dis} = -\mathbb{E}_z [ D(g_{G} (z, y_{t})) ]  - \log f_{ac} (y_{t}| g_{G} (z, y_{t})) + \\ \lambda_{gp}\mathbb{E}_{\hat{x}} [ (\left \| \bigtriangledown_{\hat{x}}D(\hat{x} ^2)  \right \|_{2}-1 )^2 ]
\end{multline}
where $f_{ac}$ represents the auxiliary classifier in the discriminator, and $\lambda_{gp}$ is the weighting parameter from \cite{GulrajaniAADC17}. The discriminative loss is the standard Wasstein loss \cite{arjovsky2017wasserstein} with the auxiliary classifier for generator.

For the outlier loss,
\begin{equation}
\setlength{\abovedisplayskip}{3pt}
\setlength{\belowdisplayskip}{3pt}
   \mathcal{L}_{out}= \frac{1}{|g(z, y_{t})|} \sum_{p_i\in g(z)} \max_{q_j \in N(p_i)}|q_j-p_i|
\end{equation}

The outlier loss calculates the maximum neighbor distance of each point, the loss encourages the generator to avoid generating points that are away from other points. A realistic point cloud is composed of dense and distortionless points, the outlier loss will prevent the generation of meaningless points.

For the uniform loss,

\begin{equation}
\setlength{\abovedisplayskip}{3pt}
\setlength{\belowdisplayskip}{3pt}
   \mathcal{L}_{UL}=\sum_i L_{imbalance} (C_{i})\cdot L_{clutter}(C_{i})
\end{equation}

To generate a more natural adversarial example, we also adopt the uniform loss \cite{li2019pu}. The uniform loss pushes the generator to generate a more uniform point cloud. The $L_{imbalance}$ tries to make each generated point cluster $C$ (clustered by distance) has the same uniform number of points. The $L_{clutter}$ encourages the generator to generate points that have expected point-to-neighbor distance.

\begin{figure}[t] 
\setlength{\abovecaptionskip}{-0.2cm}   

\setlength{\belowcaptionskip}{-0.6cm} 
   \begin{center}
     \includegraphics[width=0.6\linewidth]{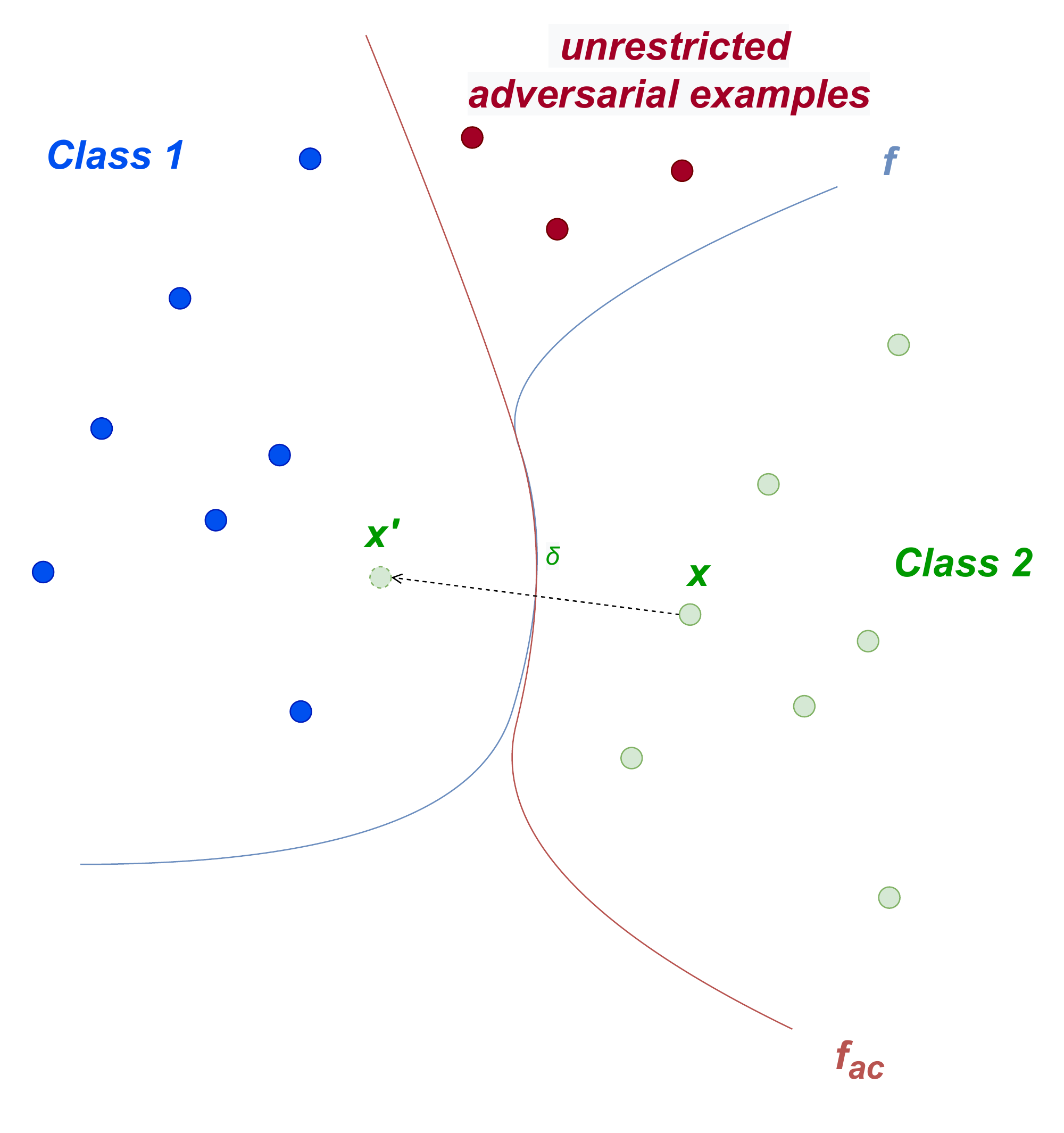}
   \end{center}
      \caption{\textbf{The existence of unrestricted adversarial examples.} Red points are the unrestricted adversarial examples, which make the target network misclassify data from class 2 into class 1. Green dotted point is the restricted adversarial example.}
   \label{fig:analysis}
   \end{figure}
   
\subsubsection{Training the AdvGCGAN}

To effectively train the proposed AdvGCGAN, the training process is divided into two stages, GAN training and adversarial training. When finished training, the AdvGCGAN is able to generate realistic adversarial examples only with random noise and the target label.

\textbf{GAN Training}: The structure of the proposed AdvGCGAN is an ACWGAN-GP network. In this stage, we aim to train the network to be able to generate a natural point cloud. We set the loss function of the generator to $\mathcal{L} = \mathcal{L}_{dis}+\lambda_2\mathcal{L}_{out}+\lambda_3\mathcal{L}_{UL}$, which means the generator keeps generating more and more realistic point clouds that can fool the discriminator. After this stage, the generator is able to generate any 3D point cloud object with a given label.

\textbf{Adversarial Training}: To successfully attack the target network, the AdvGCGAN is then trained with the adversarial attack loss, i.e. $\mathcal{L} = \mathcal{L}_{dis} + \lambda_1\mathcal{L}_{obj}+\lambda_2\mathcal{L}_{out}+\lambda_3\mathcal{L}_{UL}$. In the objective loss, we only sum the losses that fail to attack the target network. The quality of the generated point clouds can be affected if we try to train the generator to generate the point clouds with high target label confidence. Our method is doing a 3D point cloud generation task rather than being limited to specific 3D classification tasks. Therefore, in this paper, the proposed AdvGCGAN can conduct unrestricted adversarial attacks with a high success rate and high visual quality.

\subsubsection{Analysis}

We give trivial analysis about why AdvGCGAN can generate realistic adversarial examples with a GAN. 
We assume that the target network $f(x)$ is a classifier for two classes, where $x$ is a 3D point cloud data. The auxiliary classifier adopted in AdvGCGAN is $f_{ac}$. Following the previous explanation for the restricted adversarial examples \cite{goodfellow2014explaining}, the decision boundary of $f$ is nearly linear. The restricted adversarial examples are generated by finding the small enough $\delta$ that fool the target network $f$. As shown in Figure 3, $x'$ is an example of restricted adversarial examples. Suppose $f\approx w^Tx+b$, we can formulate the decision boundary of$f_{ac}$ as $f_{ac}\approx w^T_{ac}x+b_{ac}$. These two boundaries must intersect because $f$ and $f_{ac}$ are different in terms of network architecture and training. Therefore, there are gaps between these two boundaries, which are where we can find our unrestricted adversarial examples (red points in Figure 3).

\begin{table*}[ht]

\setlength{\belowcaptionskip}{-0.3cm} 
\setlength{\tabcolsep}{7.7mm}{
\begin{tabular}{c|c|c|cc}
\hline
\textbf{Class}         & \textbf{Method}                           & \textbf{\begin{tabular}[c]{@{}c@{}}Attack\\ Success Rate(\%)\end{tabular}} & \textbf{\begin{tabular}[c]{@{}c@{}}Defense\\ (SRS)\end{tabular}} & \textbf{\begin{tabular}[c]{@{}c@{}}Defense\\ (SOR)\end{tabular}} \\ \hline
\multirow{9}{*}{Chair} & IFGM                                      & 66.14                                                                       & 3.12                                                            & 2.53                                                             \\
                       & C\&W                                      & 100.0                                                                       & 0.41                                                            & 0.31                                                            \\
                       & LG-GAN                                    & 77.57                                                                       & 74.35                                                            & 76.54                                                            \\ \cline{2-5} 
                       & AdvGCGAN-PointNet                         & 94.34                                                                       & 82.83                                                            & 81.07                                                            \\
                       & \textit{AdvGCGAN-PointNet++}              & \textit{99.97}                                                              & \textit{95.96}                                                   & \textit{35.39}                                                   \\
                       & AdvGCGAN-DGCNN                            & 98.58                                                                       & 96.95                                                            & \textbf{99.97}                                                   \\
                       & AdvGCGAN$^{UL}$-PointNet & 93.64                                                                       & 81.63                                                            & 87.32                                                            \\
                       & \textit{AdvGCGAN$^{UL}$-PointNet++}              & \textit{99.97}                                                              & \textit{93.48}                                                   & \textit{32.27}                                                   \\
                       & AdvGCGAN$^{UL}$-DGCNN                            & \textbf{99.03}                                                              & \textbf{99.52}                                                   & \textbf{99.97}                                                   \\ \hline
\end{tabular}
}

      \caption{\textbf{Attack success rate (\%) of different attack methods.} The target network of IFGM, C\&W, and LG-GAN is PointNet according to the original paper. Our method's target network is given in the name of the method, and AdvGCGAN$^{UL}$ represent that the uniform loss is added in the training process.}
      
\end{table*}

\section{Quantitative Measurement}

In previous works, perturbation-based distances are commonly adopted in adversarial attack quantitative measurement, like $l_2$ norm and $\mathit{Chamfer}$ distance. However, the unrestricted adversarial examples do not have corresponding ground truth. Thus, we adopt the conventional metrics \cite{achlioptas2018learning} used in the 3D point cloud GAN network to evaluate the performance of AdvGCGAN. The reason is that we want AdvGCGAN to generate more natural and realistic point clouds that can fool both the target network and human eyes. These metrics can prove that our generated adversarial examples are similar to the real samples.

\section{Experiments}

In this section, we perform experiments to evaluate the performance of the proposed AdvGCGAN. We give comparisons among state-of-art attack methods in direct attacks and transfer attacks to test the performance and generalization of the proposed method. We also show the visualization results of different attack methods. More experiment results are attached in the supplementary materials.


\textbf{Dataset and evaluation metrics}: We adopted ShapeNet \cite{chang2015shapenet} for performance evaluation, which contains 16 object classes in total with 3D shapes. We sampled 2048 points for training the AdvGCGAN. Three categories were selected for the main performance evaluation: \textit{airplane} (2690 objects), \textit{car} (1824 objects) and \textit{chair} (3746 objects). For performance evaluation, we evaluated the attack success rate of the state-of-art attack methods for measuring the attack performance. We evaluated the JSD, MMD, and COV scores \cite{achlioptas2018learning} of the attack methods for measuring the quality of the generated adversarial examples. These scores are commonly adopted in 3D GAN for performance measurement.

\textbf{Comparisons}: We compared our method with different attack methods. IFGM \cite{liu2019extending} is an iterative gradient-based attack method, which adds perturbations over the clean point clouds for attacking the target network. Its performance on ShapeNet dataset is unsatisfying, thus we do not further discuss it in the following experiments.  C\&W \cite{xiang2019generating} uses an optimization-based algorithm that iteratively searches for the success adversarial examples. LG-GAN \cite{zhou2020lg} is the first GAN-based attack method for the 3D deep learning network. It generates the adversarial examples taking the clean point clouds as input, it adopts the adversarial loss of the target network for training the LG-GAN. 

\textbf{Performance on PointNet++ \cite{qi2017pointnet++}}: During the experiment, we found our method can directly achieve around 99.9\% attack success rate on PointNet++ without adversarial training (stage 2). The reason could be PointNet++ excessively treats the local feature, the generated point clouds are different from the original dataset in local details.  Thus, we use italic font to represent the result on PointNet++. Some attack performance can be affected because of the lack of adversarial training. 

\subsection{Attack Performance}

\begin{table*}[ht]

\setlength{\belowcaptionskip}{-0.5cm} 
\resizebox{\linewidth}{18mm}{
\begin{tabular}{l|c|cc|cc|c}
\hline
\multicolumn{1}{c|}{\textbf{Class}} & \textbf{Method}                           & \textbf{MMD-CD$\downarrow$} & \textbf{MMD-EMD$\downarrow$} & \textbf{COV-CD$\uparrow$} & \textbf{COV-EMD$\uparrow$} & \textbf{JSD$\downarrow$} \\ \hline
\multirow{7}{*}{Chair}              & AdvGCGAN                                  & 0.0023          & 0.09             & 15              & 7                & 17           \\ \cline{2-7} 
                                    & AdvGCGAN-PointNet                         & 0.0029          & \textbf{0.16}    & 3      & \textbf{6}       & 41           \\
                                    & \textit{AdvGCGAN-PointNet++}              & \textit{0.0029} & \textit{0.16}    & \textit{3}      & \textit{12}      & \textit{41}  \\
                                    & AdvGCGAN-DGCNN                            & \textbf{0.0018} & \textbf{0.16}    & \textbf{15}     & 2                & 29           \\
                                    & AdvGCGAN$^{UL}$-PointNet & 0.0031          & 0.17             & 3               & 5                & 36           \\
                                    & \textit{AdvGCGAN$^{UL}$-PointNet++}              & \textit{0.0019} & \textit{0.09}    & \textit{3}      & \textit{7}       & \textit{15}  \\
                                    & AdvGCGAN$^{UL}$-DGCNN                            & 0.0019          & \textbf{0.16}    & \textbf{13}     & 2                & \textbf{28}  \\ \hline
\end{tabular}
}
      \caption{\textbf{Generation quality of our generative method.} The results of COV-CD, COV-EMD, and JSD are multiplied by 10$^2$. Lower MMD-CD, MMD-EMD, and JSD, and Higher COV-CD and COV-EMD represent better results.}
      
\end{table*}

\begin{table}[ht]

\setlength{\belowcaptionskip}{-0.5cm} 
\resizebox{\linewidth}{15mm}{
\begin{tabular}{c|ccc}
\hline
\textbf{Method}                           & \textbf{PointNet} & \textbf{PointNet++} & \textbf{DGCNN} \\ \hline
C\&W                                      & /             & 2.15               & 1.25          \\
LG-GAN                                    & /             & 31.75               & 27.62          \\ \hline
Ours-PointNet                         & /                 & \textbf{99.98}      & 60.49          \\
Ours-DGCNN                            & 2.13              & 97.70               & \textbf{/}     \\
Ours$^{UL}$-PointNet & /                 & 99.75               & \textbf{74.74} \\
Ours$^{UL}$-DGCNN                            & 3.68              & 96.45               & \textbf{/}     \\ \hline
\end{tabular}
}
\caption{\textbf{Transfer attack success rate of different attack methods.} The original target network is marked with “/”.}

\end{table}

The attack performance evaluations are summarized in Table 1 and 2. In Table 1, we calculate the attack success rate with different classes under different defense methods. Because our attack method generates adversarial examples from scratch and does not need the input of the clean point clouds, it is totally different from the generation of the previous attack methods. We give comparisons with the previous attack methods for \textbf{reference} only.  Among all of these attack methods, the IFGM has the lowest attack success rate while the proposed AdvGCGAN outperforms by at least 20\%. The C\&W, LG-GAN, and AdvGCGAN achieve similar attack success rates when performing attacks on PointNet. We achieve a significantly higher attack success rate (almost 100\% on chair dataset) on PointNet++ and DGCNN than LG-GAN. The experiments show that our method is capable of generating powerful adversarial examples in both global and especially local features. 

We also conduct experiments attacking different defense methods \cite{zhou2019dup}: Simple Random Sampling (SRS) and Statistical Outlier Removal (SOR). In SRS, we randomly drop 20\% points from the input. In SOR, we use the same parameter as in \cite{zhou2019dup}. The result shows that our method can easily break through the defense especially when targeting DGCNN. It is slightly harder to attack PointNet with SRS because the realistic adversarial point clouds are similar in global shape to the clean point clouds. 

We evaluate the performance of AdvGCGAN with metric used in \cite{achlioptas2018learning} for standard 3D point cloud generation task. Table 2 shows that the stage 2 adversarial training will not notably decrease the generative performance of the original GAN with proper parameters. In other words, our method will generate realistic point clouds like the 3D point cloud GAN does. The performance of AdvGCGAN is sightly better when DGCNN is selected as the target network. We want to argue that with a stronger 3D generator, the GAN will generate more natural adversarial examples with adversarial training. Our two-stages training can be adapted to any 3D GAN framework with little modification. 

\subsection{Ablation Study}

\begin{figure}[t] 

\setlength{\abovecaptionskip}{-0.2cm}   

\setlength{\belowcaptionskip}{-0.5cm} 
   \begin{center}
     \includegraphics[width=0.6\linewidth]{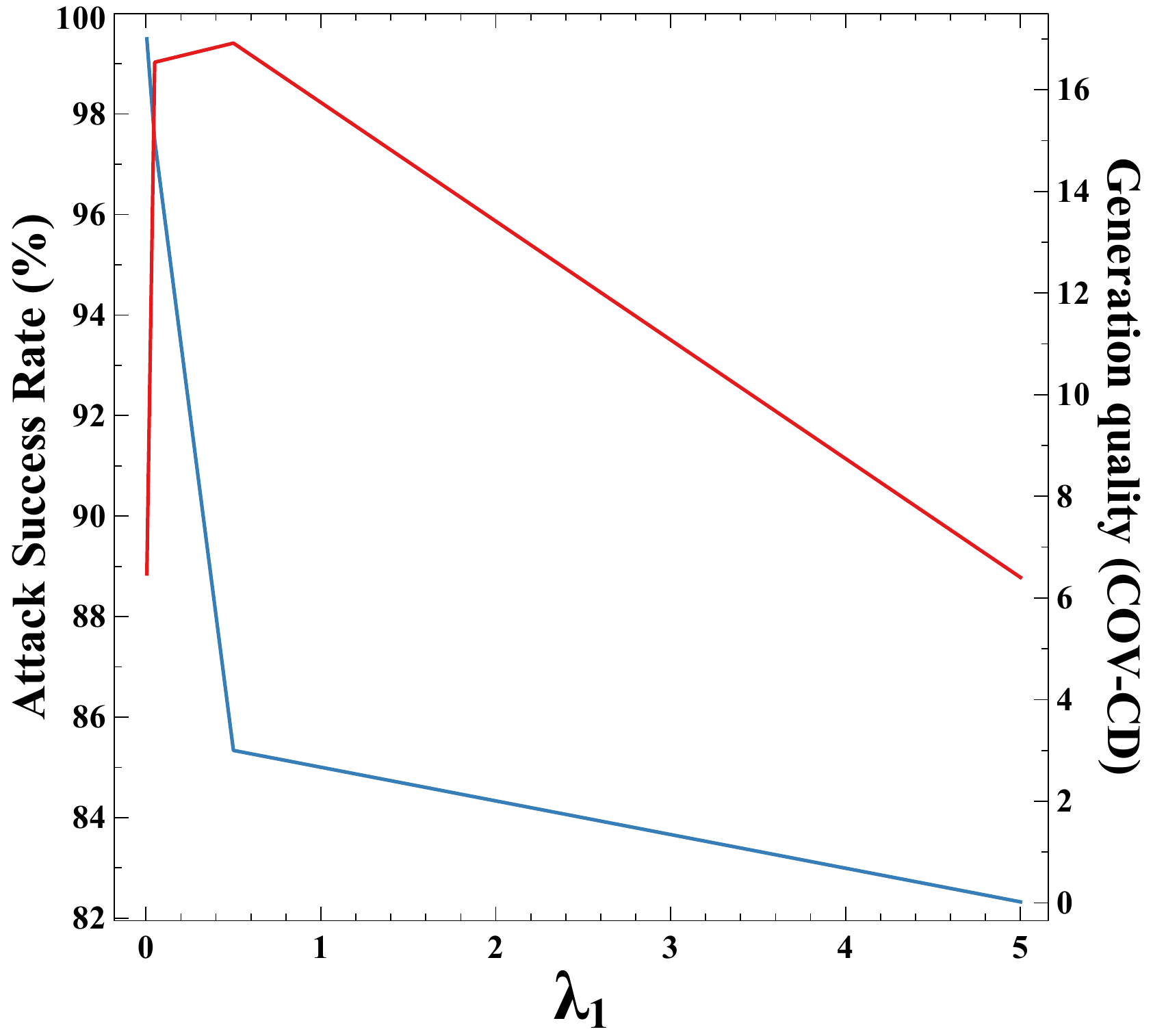}
   \end{center}
      \caption{\textbf{The effect of $\lambda_1$.} An appropriate setting of $\lambda_1$ can generate the adversarial examples with both high attack success rate and generation quality. The loss of generation quality is acceptable when $\lambda_1$ is set around 0.01.}
   \label{fig:lambda}
   \end{figure}

\textbf{Generation quality and attack success rate}: The weight of adversarial generator training can be affected by the parameter $\lambda_1$. The attack success rate will increase when $\lambda_1$ is larger, however, the generation quality will decrease as shown in Figure 4. Because the generator searches the latent space between the decision boundary of the target network. If the adversarial loss is much larger, it pushes the generator to find stronger adversarial examples rather than more realistic point clouds. However, if $\lambda_1$ is too large, the training of GAN is not stable, and convergence failure could happen. The $\lambda_2$ and $\lambda_3$ do not significantly affect the performance of the AdvGCGAN, which are discussed in the supplementary materials.

The proposed AdvGCGAN generates adversarial examples from scratch by taking a noise vector as the input. Thus, the generation quality largely depends on the generator adopted by the AdvGCGAN. We adopt the state-of-art 3D point cloud GAN, and we believe with a better 3D generator the attack performance of our method will be better. The 3D unrestricted adversarial attack is more like an general attack framework rather than being limited to a single model. 

\textbf{Training with adversarial loss}: Adversarial loss is added to train the GAN network to be able to generate adversarial examples. Unlike the LG-GAN, the AdvGCGAN generates the adversarial examples only with the noise vector. That is to say, we use the adversarial loss to train a standard GAN task for the adversarial attack. The adversarial loss will encourage the generator to generate the point clouds that make the target network give the wrong prediction while the discriminator gives the correct one. We also find the generator generates a more natural point cloud with a more accurate target network. The effect of adversarial loss can be minimized by setting a proper coefficient and giving a more strong target network like DGCNN.

\textbf{Transfer attack}: We perform transfer attack tasks with different settings. The result is shown in Table 3. our method achieves notably better than other attack methods in attack transferability, especially when training with PointNet. PointNet++ and DGCNN have similar network structures. Thus, DGCNN's adversarial examples can easily fool the PointNet++ classifier. However, PointNet takes the global information to classify the input. Therefore, DGCNN's adversarial examples do not achieve good transferability on PointNet. But PointNet's adversarial examples perform satisfying transfer attack success rates on both PointNet++ and DGCNN. The reason is that PointNet++ and DGCNN also use global information to classify the 3D point clouds. Our attack is more ''universally`` when training with the adversarial loss against PointNet. The reason for such a good transferability is that our method benefits from generating the
adversarial examples from scratch instead of adding perturbations to the original point clouds. The perturbations are more target-network-biased and limited to the norm distance to the input point clouds.
   
\subsection{Discussion}

\textbf{Qualitative results}: We give visual adversarial examples from C\&W, LG-GAN, and AdvGCGAN on two classes (\textit{airplane} and \textit{chair}). Figure 5 shows that C\&W generates many outlier points while LG-GAN generates deformed point clouds. However, the unrestricted adversarial examples generated by our AdvGCGAN are visually more natural and realistic. But the quality is somehow influenced by the adversarial loss, the deformation still occurs in some cases. The 3D point cloud generation task is still challenging, we aim to further improve it in the future study.

\textbf{Unrestricted adversarial examples}: The restricted adversarial examples from previous works mostly directly adding perturbations over the 3D point clouds, which will cause the changes of the coordinates. Even with little changes, it can affect the visualization of the point clouds. Therefore, we aim to generate point clouds without norm restriction, the proposed method directly generates adversarial examples by inputting the noise vector. Our method achieves satisfying results without fine-tuning the generator and the discriminator, which proves the effectiveness of unrestricted adversarial examples.

\begin{figure}[t] 

\setlength{\abovecaptionskip}{-0.2cm}   

\setlength{\belowcaptionskip}{-0.5cm} 
   \begin{center}
     \includegraphics[width=1.0\linewidth]{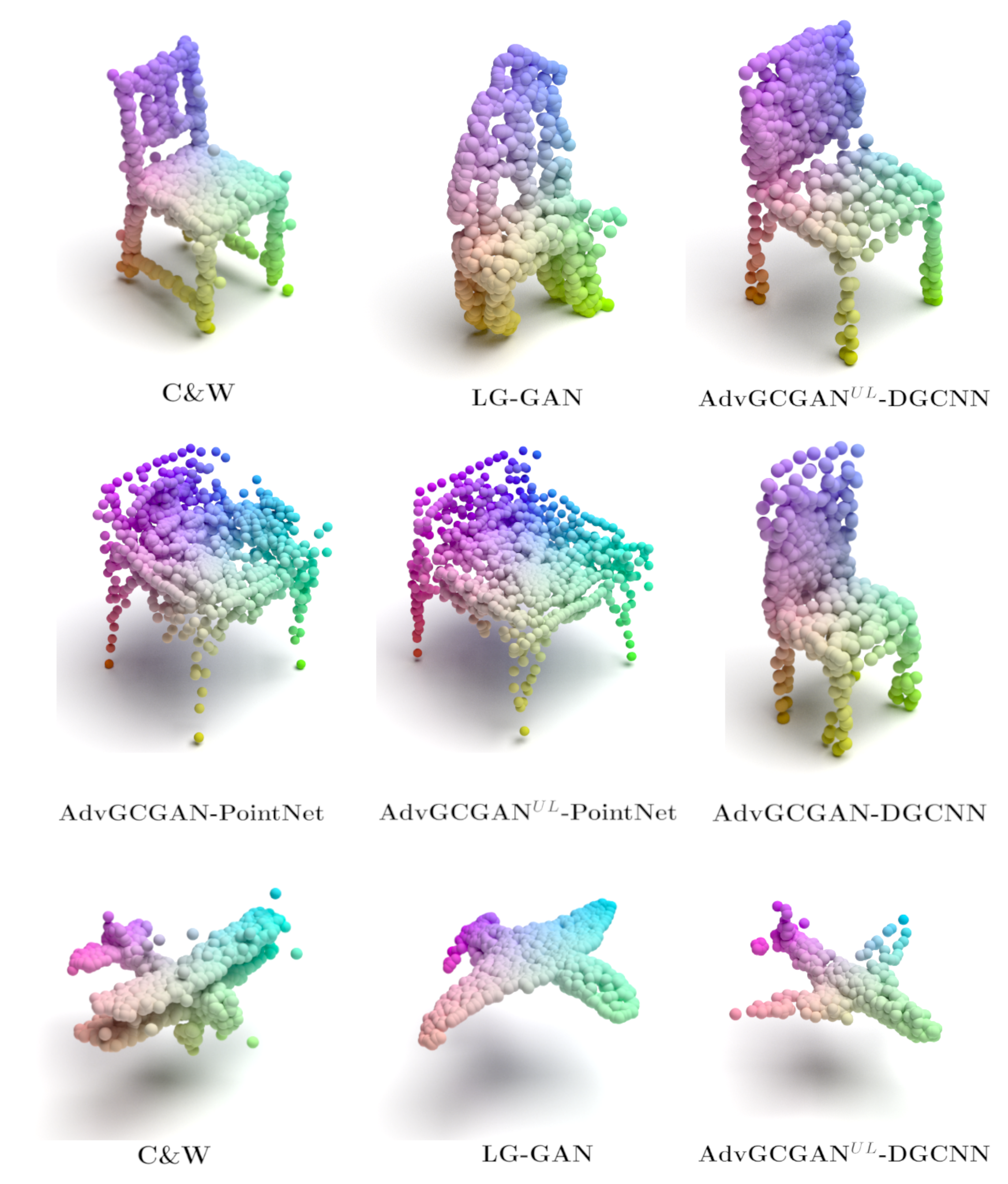}
   \end{center}
      \caption{\textbf{Qualitative results on \textit{airplane} and \textit{chair} class.} The top two rows are results from \textit{chair} class.}
   \label{fig:results}
   \end{figure}

\section{Conclusion}

In this work, we propose a 3D unrestricted adversarial examples generation network AdvGCGAN. We are the first to generate 3D adversarial examples by taking only noise vector and label as input. The specially designed GAN network architecture is utilized to generate a more realistic and natural adversarial point cloud. The two stages of training with unrestricted adversarial attack loss make the AdvGCGAN be able to generate adversarial examples that have a high attack success rate and still can fool humans. The experiments show that the proposed AdvGCGAN can successfully attack the 3D deep learning model while outperforming the state-of-art perturbation-based methods in terms of visual quality. We also find that the unrestricted adversarial examples have good transferability and can achieve satisfying performance against defense methods, which further improves the usability of our work.

{\small
\bibliographystyle{ieee_fullname}
\bibliography{egbib}
}

\end{document}